\documentclass{article}
    \PassOptionsToPackage{numbers, sort&compress}{natbib}

    \usepackage[final]{neurips_2020}

\usepackage[utf8]{inputenc} 
\usepackage[english]{babel}
\usepackage[T1]{fontenc}    
\usepackage{hyperref}       
\usepackage{url}            
\usepackage{booktabs}       
\usepackage{amsfonts}       
\usepackage{nicefrac}       
\usepackage{microtype}      
\usepackage{color}
\usepackage{graphicx}
\usepackage{caption}
\usepackage{subcaption}
\graphicspath{ {./images/} }
\usepackage{bm}
\usepackage{bbm}
\usepackage{multirow}
\usepackage{amssymb}

\title{Cross-Domain Sentiment Classification with In-Domain Contrastive Learning}
\author{%
  Tian Li\thanks{Equal Contribution} \\
  Peking University\\
  \texttt{davidli@pku.edu.cn} \\
  \And
  Xiang Chen\footnotemark[1]  \\
  Peking University \\
  \texttt{caspar@pku.edu.cn} \\
  \AND
  Shanghang Zhang\thanks{Correspondence Author}  \\
  UC Berkeley  \\
  \texttt{shz@eecs.berkeley.edu} \\
  \And
  Zhen Dong\footnotemark[2] \\
  UC Berkeley \\
  \texttt{zhendong@berkeley.edu} \\
  \And 
  Kurt Keutzer \\
  UC Berkeley \\
  \texttt{keutzer@berkeley.edu} \\
}

\begin{document}

\maketitle
\begin{abstract}
  Contrastive learning (CL) has been successful as a powerful representation learning method. In this paper, we propose a contrastive learning framework for cross-domain sentiment classification. We aim to induce domain invariant optimal classifiers rather than distribution matching. To this end, we introduce in-domain contrastive learning and entropy minimization. Also, we find through ablation studies that these two techniques behaviour differently in case of large label distribution shift and conclude that the best practice is to choose one of them adaptively according to label distribution shift. The new state-of-the-art results our model achieves on standard benchmarks show the efficacy of the proposed method. 
\end{abstract}

\section{Introduction}
Domain shift is common in language applications. One is more likely to find "internet" or "PC" in reviews on electronics than those on books, while he or she is more likely to find "writing" or "B.C." in reviews on books than those on electronics. This proposes a fundamental challenge to NLP in that many computational models fail to maintain comparable level of performance across domains. Formally, a distribution shift happens when a model is trained on data from one distribution (source domain), but the goal is to make good predictions on some other distribution (target domain) that shares the label space with the source. 

We study unsupervised domain adaptation in this work, where we have fully-labeled data on source domain but no labeled data on target domain. The most prevailing methods in this field aim to learn domain-invariant feature by aligning the source and target domains in the feature space. The pioneering works in this field try to bridge domain gap with discrepancy-based approach. ~\cite{Sejd2013equi} first introduce MMD to measure domain discrepancy in feature space and~\cite{Long2015Learning} use its variant MK-MMD as an objective to minimize domain shift. Another line of work~\cite{ganin2016domainadversarial} introduces a domain classifier and adversarial training to induce domain invariant feature, followed by works using generative models to enhance adversarial training~\cite{hoffman2018cycada}. However, note that both MMD-based approach and adversarial training formulates with a minimax optimization procedure that is widely known as hard to converge to a satisfactory local optimum~\cite{fedus2017many, duchi2016local}. Moreover, some recent works~\cite{combes2020domain, li2020rethinking, zhao2019learning} have discovered that both of them don't guarantee good adaptation and will introduce inevitable error on target domain under label distribution shift because they may render incorrect distribution matching. For example, thinking of a binary classification task, the source domain has 50\% of positive samples and 50\% of negative samples while the target domain has 30\% postive and 70\% negative. Successfully aligning these distributions in representation space requires the classifier to predict the same fraction of positive and negative on source and target. If one achieves 100\% accuracy on the source, then target accuracy will be at most 80\%, that is 20\% error at best. 


Self-supervised representation learning could be a good workaround for this problem because it enforces predictive behaviour matching~\cite{li2020rethinking} instead of distribution matching. The main idea is to learn discriminative representation that is able to genenralize across domains. ~\cite{yujiang2016learning,ziser-reichart-2018-pivot,ziser2017neural,ziser-reichart-2019-task} use sentiment-indicating pivot prediction as their auxiliary task for cross-domain sentiment analysis. The method proposed in this paper adopts contrastive learning to extract generalizable discriminative feature. Contrastive learning is a subclass of self-supervised learning that is gaining popularity thanks to recent progress~\cite{he2019moco, chen2020mocov2, chen2020simCLR, chen2020simCLRv2}. It utilizes positive and negative samples to form contrast against the queried sample on pretext tasks in order to learn meaningful representations. However, the pretext tasks must be carefully chosen.~\cite{sun2019unsupervised} shows with experiments on computer vision tasks that the transfer performance will suffer under improper pretext tasks like pixel reconstruction. 

Therefore, in this paper we explore two classic data augmentation methods in natural language processing—synonym substitution and back translation to define our pretext task. Experiments on two cross-domain sentiment classification benchmarks show the efficacy of the proposed method. We also examine whether in-domain contrastive learning and entropy minimization~\cite{wang2020fully} helps cross-domain sentiment classification under varied label distribution settings. Our main contributions in this work 
are summarized as follows:

\begin{itemize}
    \item We are the first to introduce contrastive learning to domain adaptation for NLP to the best of our knowledge.
    \item We introduce in-domain contrastive learning and entropy minimization for cross-domain sentiment classification and adaptively apply them according to label distribution shift across domains for best accuracy performance. 
    \item We carefully choose pretext tasks for contrastive learning to promote generalizable discriminative representation learning.
    \item We beat strong baselines on standard benchmarks of domain adaptation for sentiment analysis. 
\end{itemize}

\section{Related Work}
\paragraph{Cross-Domain Sentiment Classification} The most prevailing methods for unsupervised domain adaptation for sentiment classification aims to learn domain-invariant feature by aligning the source and target domains in feature space. One line of work in this field derives from~\cite{Long2015Learning}, using MMD and its variants to measure and minimize domain discrepancy\cite{sun2015return, he2018adaptive}. Another line of work follows ~\cite{ganin2016domainadversarial}, using a domain classifier and adversarial training to induce domain invariant feature~\cite{clinchant-etal-2016-domain, li2018hierarchical, ghosal2020kingdom}. However, these methods fail to take care of label shift across domains. This can cause undesired performance degradation on target domain according to our analysis in the introduction section. Another important line of work follows Structure Correspondence Learning~\cite{blitzer2006domain}. They use pivot prediction as an auxiliary task to help extract domain-invariant knowledge~\cite{pan2010cross, yujiang2016learning, ziser2017neural, ziser-reichart-2018-pivot, ziser-reichart-2019-task}. Since transformer-based language models catch on,~\cite{du2020adversarial} designs novel self-supervised "post-training" tasks for BERT~\cite{devlin2019bert} along with domain adversarial training to help domain transfer. Please note that although we also use BERT as feature extractor, we're different from it in multiple aspects: We not only use contrastive learning instead of BERT-pretraining-style self-supervised learning as is in ~\cite{du2020adversarial}, but we also get rid of distribution matching with domain adversarial training. Our competitive performance on benchmarks further illustrates the efficacy of our model.
\paragraph{Contrastive Learning} Recent developments on contrastive learning~\cite{he2019moco, chen2020mocov2, chen2020simCLR, chen2020simCLRv2} have achieved promising results on the standard representation learning benchmarks on computer vision tasks. Although there have been several works applying contrastive learning to NLP, most of them~\cite{giorgi2020declutr, fang2020cert, yang-etal-2019-reducing, klein2020contrastive, wei2020learning} concentrate on single-domain tasks like image caption retrieval, machine translation and those on the GLUE benchmark. To the best of our knowledge, we first adopt contrastive learning as an approach to facilitate domain adaptation in natural language processing.

\section{Method}
\subsection{Contrastive Learning Framework}
\begin{figure}
    \centering
    \includegraphics[width=0.8\textwidth]{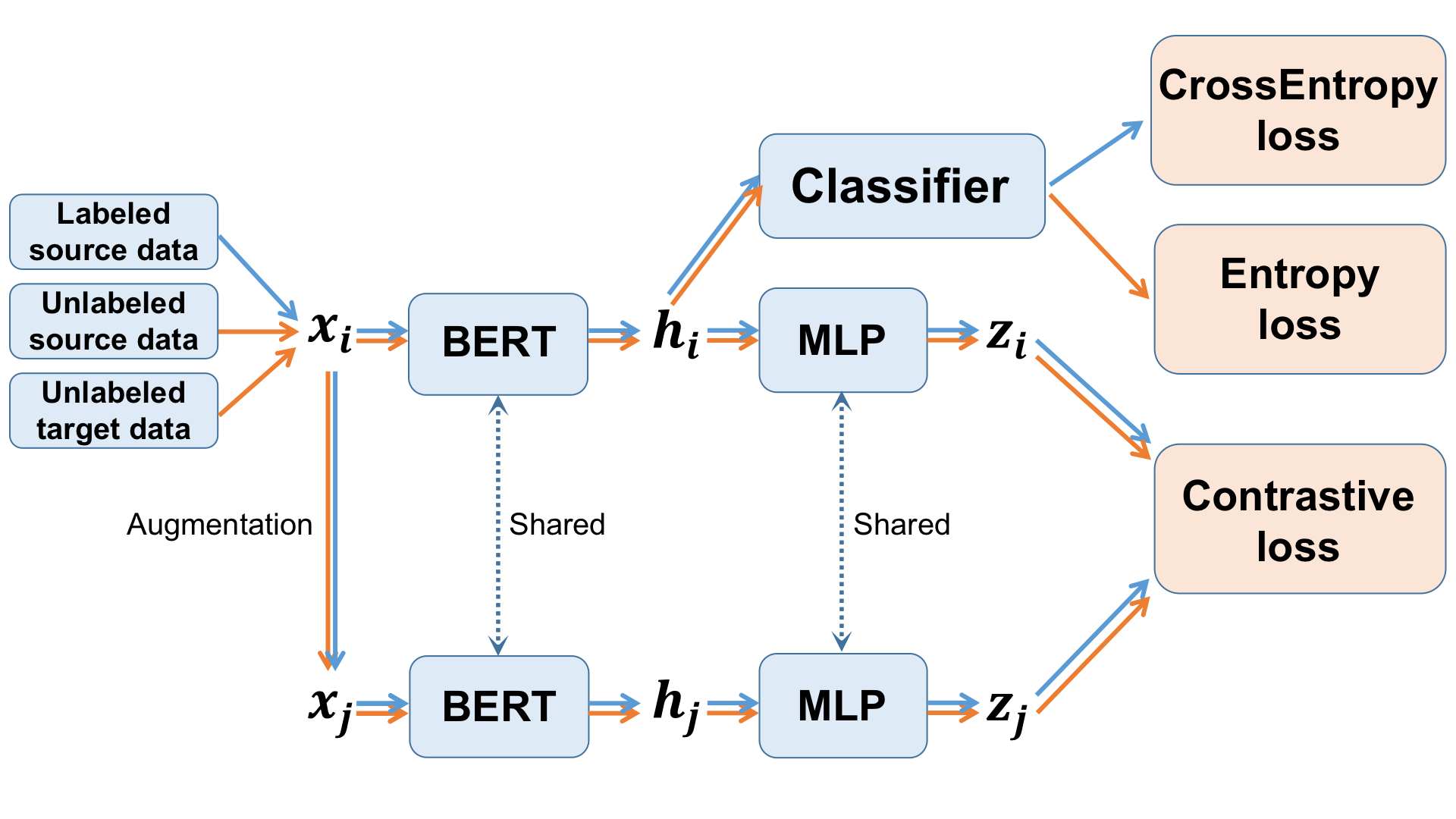}
    \caption{The whole pipeline of the proposed method: for each input document $x_i$, we first generate its positive sample $x_j$ with augmentation. $x_i$ and $x_j$ are then forwarded to the shared feature extractor BERT to get the hidden features $h_i, h_j$ 
    and a shared MLP projection head is applied subsequently to get the projected representations $z_i, z_j$. The projecion, along with that of other samples in the minibatch as negatives (omitted here for simplicity), are used to compute the contrastive loss. To induce more discriminative feature, the hidden feature are fed into the sentiment classifier to get the logits. Finally, for the unlabeled instances, they are used to compute the \emph{entropy}, while for the labeled instances, they are used to compute the \emph{cross entropy loss}. The model is jointly trained on the three objectives.}
    \label{fig:major_pipeline}
\end{figure}
We propose a framework based on~\cite{chen2020simCLR} but we elaborate it to fit the domain adaptation setting where we have unlabeled data from two domains(source and target) plus labeled data on source. As illustrated in Figure {\color{red}\ref{fig:major_pipeline}}, our framework consists of the following components:
\begin{itemize}
    \item \textbf{Positive sample generation:} We explore two classic data augmentation methods in natural language processing: synonym substitution and back translation, respectively, to generate positive sample $x_j$ for each document $x_i$.
    \item \textbf{Feature extractor:} We use pretrained BERT\cite{devlin2019bert} as our feature extractor because of its remarkable accomplishments in language understanding. Note that it's shared among all data, including their augmentations. It computes on $x_i, x_j$ independently and outputs hidden features $h_i, h_j$.
    \item \textbf{Projection head:} Following SimCLR~\cite{chen2020simCLR}, we adopt a MLP with one hidden layer applying on $h_i, {h}_j$ to get the projected representations $z_i, {z}_j$ respectively. We find that this MLP projection benefits our model in terms of learning better discriminative feature with the contrastive loss, as is found in SimCLR~\cite{chen2020simCLR}.
    \item \textbf{Contrastive loss:} We introduce in-domain contrastive loss based on the InfoNCE loss~\cite{chen2020simCLR} in this work. Details are discussed in section {\color{red} \ref{sec:contrastive_loss}}.
    \item \textbf{Sentiment classifier:} We use another one-hidden-layer MLP as our sentiment classifier to generate the predictions. The logits are then used to compute the information entropy as the \emph{entropy loss} for the unlabeled instances, while for the labeled instances, they are used to compute the ordinary cross entropy loss w.r.t the ground truth sentiment labels. Intuitions for the \emph{entropy loss} are discussed in section {\color{red} \ref{sec:entropy_loss}}.
\end{itemize}

\subsection{In-Domain Contrastive Loss}
\label{sec:contrastive_loss}
Contrastive loss function is designed to maximize the similarities between positive pairs and minimize the similarities of negative ones. At each iteration of the training, we randomly sample a minibatch $\{\bm{x}_i\}_{i=1}^N$ of $N$ samples and generate positive pairs $\{(\bm{x}_i, \bm{x}_{i+N})\}_{i=1}^N$ for each of them. As proposed by~\cite{chen2020simCLR, giorgi2020declutr}, for each positive pairs, we treat other $2(N-1)$ samples as their negative examples. The contrastive InfoNCE loss function is defined as:
\begin{equation}
    \mathcal{L}_{con}(\bm{z})=\frac{1}{2N}\sum_{k=1}^{2N}[l(\bm{z}_{2k-1}, \bm{z}_{2k}), l(\bm{z}_{2k}, \bm{z}_{2k-1})]
\end{equation}
\begin{equation}
    l(\bm{z}_i,\bm{z}_j)=-\log\frac{\exp(\textit{sim}(\bm{z}_i, \bm{z}_j)/\tau)}{\sum_{k=1}^{2N}\mathbbm{1}_{[k\neq i]}\exp(\textit{sim}(\bm{z}_i, \bm{z}_k)/\tau)}
\end{equation}

where $\mathbbm{1}_{[k\neq i]}$ is an indicator function equaling to $0$ iff $k=i$, and $\textit{sim}(\bm{z}_i, \bm{z}_j)=\bm{z}_i^\top\bm{z}_j/(\|\bm{z}_i\|\|\bm{z}_j\|)$ denotes the cosine similarity between hidden representations $\bm{z}_i$ and $\bm{z}_j$. $\tau$ is temperature parameter.

However, we noticed that contrastive loss of this form is not suitable to be directly applied to our domain adaptation setting. If we sample a minibatch that contains two instances $\bm{x}_i$ and $\bm{x}_j$ from different domains, optimizing the contrastive loss above will even widen the distance between $\bm{z}_i$ and $\bm{z}_j$. This will back-propagate to the hidden features $\bm{h}_i$ and $\bm{h}_j$ thus enlarging the domain discrepancy in hidden space.

Therefore, we proposed to perform contrastive learning in the source and target domain independently, i.e. in-domain contrastive learning. At each iteration we randomly sample $N$ instances $\bm{x}^{(s)}=\{\bm{x}^{(s)}_i\}_{i=1}^N$ from source domain and $N$ instances $\bm{x}^{(t)}=\{\bm{x}^{(t)}_i\}_{i=1}^N$ from target domain. The in-domain contrastive loss is defined as the sum of contrastive loss from both domains:
\begin{equation}
    \mathcal{L}_{con}=\mathcal{L}_{con}(\bm{z}^{(s)})+\mathcal{L}_{con}(\bm{z}^{(t)})
\end{equation}
where $\bm{z}^{(s)},\bm{z}^{(t)}$ is the projected representation of $\bm{x}^{(s)},\bm{x}^{(t)}$. 

In section {\color{red}\ref{sec:experiment_results} }, we show that in-domain contrastive learning boosts domain transfer performance especially when the target domain has significant inbalanced label distribution.

\subsection{Entropy Minimization}
\label{sec:entropy_loss}
Note that in unsupervised domain adaptation we have labels on the source domain. Therefore it's easy to train an optimal classifier for the source domain. Despite lack of label on the target domain, we can still align the optimal classifier for the target domain with the source~\cite{ahuja2020invariant}. To this end, we minimize the entropy of the model's prediction in order to disambiguate the positive and negative instances over the unlabeled data, especially on the target domain~\cite{wang2020fully}. In this way, the margin between positive and negative clusters on target domain will be widened so that there is a larger chance that the optimal decision boundary for the source domain falls within it. 

However, our model fails when applying the entropy loss starting from the first epoch. This is probably because the model needs to get a general picture of the source and target domains with contrastive learning in the first place and draw the decision boundary for the source domain. Applying the entropy loss too hastily asks the model to early decide labels for uncertain instances before the model has learnt good representations and the classification boundary on source domain. Therefore, we apply the entropy loss from the second epoch and it works out as expected. 


\section{Experiments}
\subsection{Experiment Setting}
\label{sec:experiments}
\paragraph{Cross-domain Sentiment Classification} To demonstrate the efficacy of our model, we conduct experiments on the Amazon-review\footnote{Dataset can be found at http://www.cs.jhu.edu/~mdredze/datasets/sentiment/index2.html} dataset~\cite{blitzer-etal-2007-biographies} and Airlines dataset\footnote{Dataset and process procedures can be found at https://github.com/quankiquanki/skytrax-reviews-dataset}. We follow~\cite{ziser-reichart-2018-pivot} to introduce the Airlines dataset because the label distribution on this domain is relatively more balanced compared to the domains in the Amazon-review dataset. As is shown in Table {\color{red}\ref{tab:label_distribution}}, this dataset contains reviews for four kinds of products, books, DVD, electronics, and kitchen, each defining a domain. Each domain has 1000 sample labeled as positive and negative respectively, thus the 2000 labeled data. But the number of unlabeled data and its label distribution on each domain is different, as is shown in the last two columns of the table~\cite{ziser-reichart-2018-pivot}. Note that the ratio of positive over negative on the domains in this dataset is significantly large. However, in the airline domain there are only 1.15 positive reviews for every negative review. Therefore we follow~\cite{ziser-reichart-2018-pivot} to introduce this dataset to demonstrate our model's performance when the target domain's label distribution varies across a wide range. To align the setting with the Amazon-review dataset, we randomly sample 1000 positive review and 1000 negative review from the airlines dataset as labeled data, and the rest of data are removed of labels as unlabeled data. 

It's important to clarify that in terms of label distribution shift in this paper we care about the label distribution of \emph{labeled} data on source domain and that of \emph{unlabeled} data on target domain. This is because the optimal classifier is trained with only labeled data on source domain but we want the unlabeled data on target domain to share the same optimal classifier. Also note that the labeled data of the domains involved in the experiments of this paper all have balanced label distribution, so the label distribution shift is only determined by target domain.

\begin{table}[h]
\centering
\begin{tabular}{ |c | c| c | c | c | } 
\hline
\textbf{Domains} & \textbf{labeled} & \textbf{unlabeled} & \textbf{pos:neg} \\ 
\hline
Books & 2000 & 6000 & 6.43:1\\ 
\hline
DVD & 2000 & 34741 & 7.39:1 \\ 
\hline
Electronics & 2000 & 13153 & 3.65:1 \\
\hline
Kitchen & 2000 & 16785 & 4.61:1 \\
\hline
Airlines & 2000 & 39396 & 1.15:1 \\
\hline
\end{tabular}
\setlength{\abovecaptionskip}{3pt}
\caption{The statistics of domains on the Amazon-review dataset and the Airlines dataset reported by~\cite{ziser-reichart-2018-pivot}. "pos:neg" denotes the ratio of unlabeled positive samples over unlabeled negative samples on that domain.}
\label{tab:label_distribution}
\end{table}

\paragraph{Baselines} We mainly compare our model with the state-of-the-art method BERT-DAAT~\cite{du2020adversarial} on this benchmark. We also train BERT on the source labeled data and directly test on the target labeled data, constituting a second strong baseline called BERT-base. Results of two non-BERT methods  HATN~\cite{li2018hierarchical} and IATN~\cite{zhang2019interactive} are also included.


\paragraph{Augmentation methods} We use synonym substitution and back translation as augmentation methods to generate positives in this paper. For synonym substitution, we use the method provided by python nlpaug library~\cite{ma2019nlpaug} based on WordNet~\cite{miller1995wordnet}. For back translation, we use the method provided by nlpaug library based on fairseq~\cite{ott2019fairseq}. We do random synonym substitution online, meaning that it's different among repeated training runs. But We do back translation as offline preprocessing since it's slower. To be specific, we translate the English texts to German and then back.

\paragraph{Hyperparameter tuning} We adopt the pretrained BERT-base-uncased model from huggingface~\cite{2019arXiv191003771W} in this paper. We use ReLU as the activation function for both projection head and sentiment classifier. On the neural network training, we use the AdamW~\cite{loshchilov2018fixing} optimizer with learning rate $2e-5$, linear learning rate scheduler, linear learning rate warm up, warm up steps 0.1 of total training steps  and weight decay 0.01. We train the model for 4 epochs and set the temperature parameter of the contrastive loss to 0.05. For synonym substitution, we set the augmentation rate to 0.3 and remove max limit to the number of augmented words. For back translation, we set the beam parameter to 1.

\subsection{Experiment Results}
\label{sec:experiment_results}
\begin{table}[h]
    \centering
    \begin{tabular}{c c c | c c | c c}
    \toprule[1.5pt]
         \multirow{2}{*}{S$\rightarrow$ T} & \multicolumn{2}{c|}{Previous Models} & \multicolumn{4}{c}{BERT}\\
         \cline{2-7} & HATN & IATN & BERT-base & BERT-DAAT  & BERT-CL$^{bt}$ & BERT-CL$^{ss}$ \\\midrule[1pt]
         B$\rightarrow$ E & 85.70 & 86.50 & 90.50 & 89.57           &  \textbf{91.48} & 90.99 \\
         K$\rightarrow$ D & 84.50 & 84.10 & 87.90 & 88.81 & \textbf{88.99}          &  87.65 \\
         \emph{Average} & 85.10 & 85.30 & 89.20        &   89.19           & \textbf{90.23}         & 89.32 \\\midrule[1pt]
         B$\rightarrow$ A &--&-- & 86.18 &  --             &      \textbf{86.56}         & 83.61 \\    
    \bottomrule[1.5pt]
    \end{tabular}
    \setlength{\abovecaptionskip}{5pt}    
    \caption{Sentiment classification accuracy on the Amazon-review dataset and Airlines dataset. We only evaluate on three domain settings as an initial trial. BERT-base denotes BERT trained on source and direcly tested on target, BERT-DAAT is the method proposed in~\cite{du2020adversarial}. BERT-CL$^{bt}$ is our model with back translation as augmentation. BERT-CL$^{ss}$ is our model with synonym substitution as augmentation. B, D, E, K denotes the 4 domains in Amazon-review dataset respectively and A denotes the airlines domain.}
    \label{tab:major_results}
\end{table}
\paragraph{Comparison with baseline} Table {\color{red}\ref{tab:major_results}} shows our model's performance on the benchmarks. We find that back translation is generally better than synonym substitution as an augmentation method for contrastive learning. Although we don't beat BERT-DAAT in all settings, both of our models are able to surpass it on a average basis. Note that our strong baseline model BERT-base has comparable accuracy with BERT-DAAT on average. 

Note that we adaptively choose techniques for our model according to label distribution shift across the source and target domains. For B$\to$E and B$\to$A settings we use entropy minimization and ordinary InfoNCE loss instead of in-domain contrastive loss because the label distribution shift is relatively small. For K$\to$D setting we apply in-domain contrastive loss but not entropy minimization because the label distribution shift is larger. The intuitions for the choice are discussed at length in the ablation studies.


\begin{table}
\begin{minipage}{\textwidth}
 \begin{minipage}[t]{0.5\textwidth}
  \centering
  \setlength{\tabcolsep}{1.4mm}{
    \begin{tabular}{cccc} 
    \toprule[1.5pt]
    \textbf{S$\to$T} & \textbf{in-domain} & \textbf{entropy loss} & \textbf{accuracy} \\
    \midrule[1pt]
    \multirow{4}{*}{B$\to$E}  & & & 90.49 \\
    & \checkmark &              & 90.05 \\ 
    &            &  \checkmark & \textbf{91.48} \\ 
    &  \checkmark &  \checkmark & 91.36\\
    \bottomrule[1.5pt]
    \end{tabular}}
  \end{minipage}
  \begin{minipage}[t]{0.5\textwidth}
   \centering
   \setlength{\tabcolsep}{1.4mm}{
    \begin{tabular}{cccc} 
    \toprule[1.5pt]
    \textbf{S$\to$T} & \textbf{in-domain} & \textbf{entropy loss} & \textbf{accuracy} \\
    \midrule[1pt]
    \multirow{4}{*}{K$\to$D} & & &88.39 \\
    & \checkmark & & \textbf{88.99} \\ 
    & & \checkmark & 87.25 \\
    & \checkmark & \checkmark & 86.36\\
    \bottomrule[1.5pt]
    \end{tabular}}
   \end{minipage}
\end{minipage}
\setlength{\abovecaptionskip}{5pt}
\caption{Ablation study of in-domain contrastive learning (section {\color{red}\ref{sec:contrastive_loss}}) and entropy minimization (section {\color{red} \ref{sec:entropy_loss}}) on the Amazon-review dataset. Note that all experiments in this table use back translation as default augmentation method.}
\label{tab:ablation}
\end{table}

\paragraph{Ablation studies} Table {\color{red}\ref{tab:ablation}} shows the results of our ablation study on in-domain contrastive learning and entropy minimization. Interestingly, we find that in-domain contrastive learning gain much benefit on K$\to$D domains but doesn't help on B$\to$E domain setting. 

\begin{figure}[h]
     \centering
     \begin{subfigure}[b]{0.3\textwidth}
         \centering
         \includegraphics[width=\textwidth]{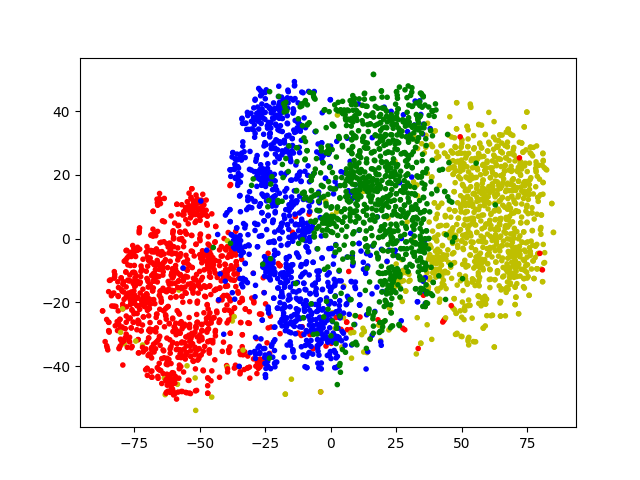}
         \caption{Baseline\\ \ }
         \label{fig:baseline}
     \end{subfigure}
     \hfill
     \begin{subfigure}[b]{0.3\textwidth}
         \centering
         \includegraphics[width=\textwidth]{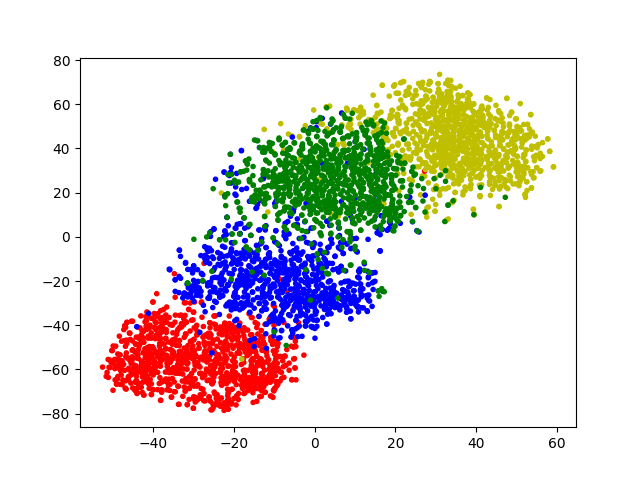}
         \caption{In-domain contrastive learning without entropy minimization}
         \label{fig:in-domain}
     \end{subfigure}
     \hfill
     \begin{subfigure}[b]{0.3\textwidth}
         \centering
         \includegraphics[width=\textwidth]{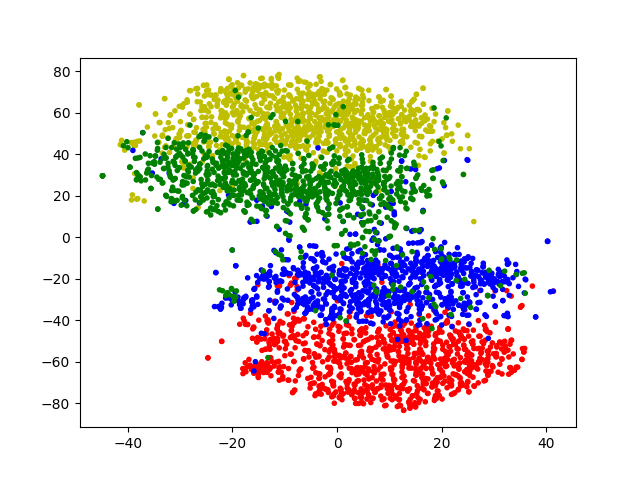}
         \caption{In-domain contrastive learning with entropy minimization}
         \label{fig:entropy}
     \end{subfigure}
        \caption{t-SNE~\cite{maaten2008visualizing} projection of (a) BERT-base hidden feature, (b) hidden feature of BERT-CL without entropy minimization, (c) hidden feature of BERT-CL full model. The red and yellow dots respectively represent the feature of positive and negative samples on the source domain, and the blue and green dots represents those on the target domain. Note that the margin between positive cluster and negative cluster on target domain becomes clearer from left to right.}
        \label{fig:TSNE_figure}
\end{figure}

In-domain contrastive learning promotes contrast between positive and negative clusters intra-domain, thus widens the decision margin and helps classification on target domain as well as source domain. On the contrary, ordinary contrastive learning additionally forms contrast inter-domain and thus increases domain discrepancy. This is better explained with visualization. Look at Figure {\color{red}\ref{fig:in-domain}}. The red and yellow dots respectively represent the feature of positive and negative samples on the source domain, and the blue and green dots represent those on the target domain. In-domain contrastive learning focuses on pushing red points away from yellow and green points away from blue, which desirably enlarge the margin between positive and negative. On the other hand, although ordinary contrastive learning does the same thing as above, it additionally pushes blue away from red and green away from yellow, which will potentially entangle the positive and negative clusters on the target domain.

However, the tradeoff here is that in-domain contrastive learning prevents the divergence of source and target domains in feature space at the cost of not fully exploiting the training data. Insufficient utilizing data will consistently cause performance loss. But the benefit brought by in-domain contrastive learning is larger as the label distribution shift across domains is larger. The underlying fact is that divergence between source and target domains will potentially push points near the decision boundary to cross it.
In particular, when the label distribution shift is significant such as 1:1 to 7.39:1 in the K$\to$D setting, there are more positive points near the decision boundary. Divergence between the kitchen and DVD domains will make instances of the positive instances on DVD domain more prone to cross the optimal decision boundary for kitchen domain. But when the label distribution shift is not so large, this effect will not be so salient. The benefit of preventing divergence is small so as to be offset by the defect of insufficient utilization of training data. That's why we find in-domain contrastive learning doesn't work well on B$\to$E setting.

In the meantime, entropy minimization also behaves differently for the B$\to$E domain and the K$\to$D domain. It performs poor on domain K$\to$D where the label distribution on the target domain is significantly unbalanced. This is in line with findings in~\cite{wang2020fully} that entropy minimization may cause the model to exploit over-represented features, i.e. it will bias to the dominant domain and the dominant class in the domain. The induced bias is likely to undermine the model's performance in case of large label distribution shift.

Given that in-domain contrastive learning and entropy minimization behaves oppositely in case of large label distribution shift, we conclude that the best practice is to use one of them adaptively according to the scale of label distribution shift. Use entropy minimization when it's small but use in-domain contrastive learning when it's large.

\paragraph{Visualization} Figure {\color{red}\ref{fig:TSNE_figure}} shows the hidden features of our model trained on the books as source and electronics as target, projected to 2D space with t-SNE~\cite{maaten2008visualizing}. Figure {\color{red}\ref{fig:baseline}} shows that BERT-base can only learn discriminative feature on source domain, while Figure {\color{red}\ref{fig:in-domain}} shows that our BERT with in-domain contrastive learning is able to learn better and approximately align the optimal decision boundary on source and target domain. Finally, Figure {\color{red}\ref{fig:entropy}} demonstrates that, further with entropy minimization, our model is able to disambiguate on the target domain and widen the gap between target positive and negative clusters.

\section{Conclusion and Broader Impact}
In this paper, we propose in-domain contrastive learning with entropy minimization to promote domain transfer. Our proposed model beats strong baselines and visualization results also show the efficacy of our model. Extensive ablation studies unveil how label distribution shift may interact with our model. It remains open question how to address label shift across domains. Besides, although synonym substitution works better than back translation in this paper, it's still interesting to explore more data augmentation methods and summarize what serve best for contrastive learning for cross-domain sentiment classification.



\bibliographystyle{plainnat}
\bibliography{reference}

\begin{thebibliography}{42}
\providecommand{\natexlab}[1]{#1}
\providecommand{\url}[1]{\texttt{#1}}
\expandafter\ifx\csname urlstyle\endcsname\relax
  \providecommand{\doi}[1]{doi: #1}\else
  \providecommand{\doi}{doi: \begingroup \urlstyle{rm}\Url}\fi

\bibitem[Ahuja et~al.(2020)Ahuja, Shanmugam, Varshney, and
  Dhurandhar]{ahuja2020invariant}
Kartik Ahuja, Karthikeyan Shanmugam, Kush~R. Varshney, and Amit Dhurandhar.
\newblock Invariant risk minimization games, 2020.

\bibitem[Blitzer et~al.(2006)Blitzer, McDonald, and Pereira]{blitzer2006domain}
John Blitzer, Ryan McDonald, and Fernando Pereira.
\newblock Domain adaptation with structural correspondence learning.
\newblock In \emph{Proceedings of the 2006 conference on empirical methods in
  natural language processing}, pages 120--128, 2006.

\bibitem[Blitzer et~al.(2007)Blitzer, Dredze, and
  Pereira]{blitzer-etal-2007-biographies}
John Blitzer, Mark Dredze, and Fernando Pereira.
\newblock Biographies, {B}ollywood, boom-boxes and blenders: Domain adaptation
  for sentiment classification.
\newblock In \emph{Proceedings of the 45th Annual Meeting of the Association of
  Computational Linguistics}, pages 440--447, Prague, Czech Republic, June
  2007. Association for Computational Linguistics.
\newblock URL \url{https://www.aclweb.org/anthology/P07-1056}.

\bibitem[Chen et~al.(2020{\natexlab{a}})Chen, Kornblith, Norouzi, and
  Hinton]{chen2020simCLR}
Ting Chen, Simon Kornblith, Mohammad Norouzi, and Geoffrey Hinton.
\newblock A simple framework for contrastive learning of visual
  representations.
\newblock \emph{arXiv preprint arXiv:2002.05709}, 2020{\natexlab{a}}.

\bibitem[Chen et~al.(2020{\natexlab{b}})Chen, Kornblith, Swersky, Norouzi, and
  Hinton]{chen2020simCLRv2}
Ting Chen, Simon Kornblith, Kevin Swersky, Mohammad Norouzi, and Geoffrey
  Hinton.
\newblock Big self-supervised models are strong semi-supervised learners.
\newblock \emph{arXiv preprint arXiv:2006.10029}, 2020{\natexlab{b}}.

\bibitem[Chen et~al.(2020{\natexlab{c}})Chen, Fan, Girshick, and
  He]{chen2020mocov2}
Xinlei Chen, Haoqi Fan, Ross Girshick, and Kaiming He.
\newblock Improved baselines with momentum contrastive learning.
\newblock \emph{arXiv preprint arXiv:2003.04297}, 2020{\natexlab{c}}.

\bibitem[Clinchant et~al.(2016)Clinchant, Csurka, and
  Chidlovskii]{clinchant-etal-2016-domain}
St{\'e}phane Clinchant, Gabriela Csurka, and Boris Chidlovskii.
\newblock A domain adaptation regularization for denoising autoencoders.
\newblock In \emph{Proceedings of the 54th Annual Meeting of the Association
  for Computational Linguistics (Volume 2: Short Papers)}, pages 26--31,
  Berlin, Germany, August 2016. Association for Computational Linguistics.
\newblock \doi{10.18653/v1/P16-2005}.
\newblock URL \url{https://www.aclweb.org/anthology/P16-2005}.

\bibitem[Combes et~al.(2020)Combes, Zhao, Wang, and Gordon]{combes2020domain}
Remi Tachet~des Combes, Han Zhao, Yu-Xiang Wang, and Geoff Gordon.
\newblock Domain adaptation with conditional distribution matching and
  generalized label shift.
\newblock \emph{arXiv preprint arXiv:2003.04475}, 2020.

\bibitem[Devlin et~al.(2019)Devlin, Chang, Lee, and Toutanova]{devlin2019bert}
Jacob Devlin, Ming-Wei Chang, Kenton Lee, and Kristina Toutanova.
\newblock Bert: Pre-training of deep bidirectional transformers for language
  understanding, 2019.

\bibitem[Du et~al.(2020)Du, Sun, Wang, Qi, and Liao]{du2020adversarial}
Chunning Du, Haifeng Sun, Jingyu Wang, Qi~Qi, and Jianxin Liao.
\newblock Adversarial and domain-aware bert for cross-domain sentiment
  analysis.
\newblock In \emph{Proceedings of the 58th Annual Meeting of the Association
  for Computational Linguistics}, pages 4019--4028, 2020.

\bibitem[Duchi et~al.(2016)Duchi, Lafferty, Zhu, et~al.]{duchi2016local}
John~C Duchi, John Lafferty, Yuancheng Zhu, et~al.
\newblock Local minimax complexity of stochastic convex optimization.
\newblock In \emph{Advances in Neural Information Processing Systems}, pages
  3423--3431, 2016.

\bibitem[Fang and Xie(2020)]{fang2020cert}
Hongchao Fang and Pengtao Xie.
\newblock Cert: Contrastive self-supervised learning for language
  understanding.
\newblock \emph{arXiv preprint arXiv:2005.12766}, 2020.

\bibitem[Fedus et~al.(2017)Fedus, Rosca, Lakshminarayanan, Dai, Mohamed, and
  Goodfellow]{fedus2017many}
William Fedus, Mihaela Rosca, Balaji Lakshminarayanan, Andrew~M Dai, Shakir
  Mohamed, and Ian Goodfellow.
\newblock Many paths to equilibrium: Gans do not need to decrease a divergence
  at every step.
\newblock \emph{arXiv preprint arXiv:1710.08446}, 2017.

\bibitem[Ganin et~al.(2016)Ganin, Ustinova, Ajakan, Germain, Larochelle,
  Laviolette, Marchand, and Lempitsky]{ganin2016domainadversarial}
Yaroslav Ganin, Evgeniya Ustinova, Hana Ajakan, Pascal Germain, Hugo
  Larochelle, François Laviolette, Mario Marchand, and Victor Lempitsky.
\newblock Domain-adversarial training of neural networks, 2016.

\bibitem[Ghosal et~al.(2020)Ghosal, Hazarika, Majumder, Roy, Poria, and
  Mihalcea]{ghosal2020kingdom}
Deepanway Ghosal, Devamanyu Hazarika, Navonil Majumder, Abhinaba Roy, Soujanya
  Poria, and Rada Mihalcea.
\newblock Kingdom: Knowledge-guided domain adaptation for sentiment analysis.
\newblock \emph{arXiv preprint arXiv:2005.00791}, 2020.

\bibitem[Giorgi et~al.(2020)Giorgi, Nitski, Bader, and Wang]{giorgi2020declutr}
John~M Giorgi, Osvald Nitski, Gary~D Bader, and Bo~Wang.
\newblock Declutr: Deep contrastive learning for unsupervised textual
  representations.
\newblock \emph{arXiv preprint arXiv:2006.03659}, 2020.

\bibitem[He et~al.(2019)He, Fan, Wu, Xie, and Girshick]{he2019moco}
Kaiming He, Haoqi Fan, Yuxin Wu, Saining Xie, and Ross Girshick.
\newblock Momentum contrast for unsupervised visual representation learning.
\newblock \emph{arXiv preprint arXiv:1911.05722}, 2019.

\bibitem[He et~al.(2018)He, Lee, Ng, and Dahlmeier]{he2018adaptive}
Ruidan He, Wee~Sun Lee, Hwee~Tou Ng, and Daniel Dahlmeier.
\newblock Adaptive semi-supervised learning for cross-domain sentiment
  classification.
\newblock \emph{arXiv preprint arXiv:1809.00530}, 2018.

\bibitem[Hoffman et~al.(2018)Hoffman, Tzeng, Park, Zhu, Isola, Saenko, Efros,
  and Darrell]{hoffman2018cycada}
Judy Hoffman, Eric Tzeng, Taesung Park, Jun-Yan Zhu, Phillip Isola, Kate
  Saenko, Alexei Efros, and Trevor Darrell.
\newblock Cycada: Cycle-consistent adversarial domain adaptation.
\newblock In \emph{International conference on machine learning}, pages
  1989--1998. PMLR, 2018.

\bibitem[Klein and Nabi(2020)]{klein2020contrastive}
Tassilo Klein and Moin Nabi.
\newblock Contrastive self-supervised learning for commonsense reasoning, 2020.

\bibitem[Li et~al.(2020)Li, Wang, Che, Zhang, Zhao, Xu, Zhou, Bengio, and
  Keutzer]{li2020rethinking}
Bo~Li, Yezhen Wang, Tong Che, Shanghang Zhang, Sicheng Zhao, Pengfei Xu, Wei
  Zhou, Yoshua Bengio, and Kurt Keutzer.
\newblock Rethinking distributional matching based domain adaptation.
\newblock \emph{arXiv preprint arXiv:2006.13352}, 2020.

\bibitem[Li et~al.(2018)Li, Wei, Zhang, and Yang]{li2018hierarchical}
Zheng Li, Ying Wei, Yu~Zhang, and Qiang Yang.
\newblock Hierarchical attention transfer network for cross-domain sentiment
  classification.
\newblock In \emph{Thirty-Second AAAI Conference on Artificial Intelligence},
  2018.

\bibitem[Long et~al.(2015)Long, Cao, Wang, and Jordan]{Long2015Learning}
Mingsheng Long, Yue Cao, Jianmin Wang, and Michael Jordan.
\newblock Learning transferable features with deep adaptation networks.
\newblock volume~37 of \emph{Proceedings of Machine Learning Research}, pages
  97--105, Lille, France, 07--09 Jul 2015. PMLR.
\newblock URL \url{http://proceedings.mlr.press/v37/long15.html}.

\bibitem[Loshchilov and Hutter(2018)]{loshchilov2018fixing}
Ilya Loshchilov and Frank Hutter.
\newblock Fixing weight decay regularization in adam.
\newblock 2018.

\bibitem[Ma(2019)]{ma2019nlpaug}
Edward Ma.
\newblock Nlp augmentation.
\newblock https://github.com/makcedward/nlpaug, 2019.

\bibitem[Maaten and Hinton(2008)]{maaten2008visualizing}
Laurens van~der Maaten and Geoffrey Hinton.
\newblock Visualizing data using t-sne.
\newblock \emph{Journal of machine learning research}, 9\penalty0
  (Nov):\penalty0 2579--2605, 2008.

\bibitem[Miller(1995)]{miller1995wordnet}
George~A Miller.
\newblock Wordnet: a lexical database for english.
\newblock \emph{Communications of the ACM}, 38\penalty0 (11):\penalty0 39--41,
  1995.

\bibitem[Ott et~al.(2019)Ott, Edunov, Baevski, Fan, Gross, Ng, Grangier, and
  Auli]{ott2019fairseq}
Myle Ott, Sergey Edunov, Alexei Baevski, Angela Fan, Sam Gross, Nathan Ng,
  David Grangier, and Michael Auli.
\newblock fairseq: A fast, extensible toolkit for sequence modeling.
\newblock In \emph{Proceedings of NAACL-HLT 2019: Demonstrations}, 2019.

\bibitem[Pan et~al.(2010)Pan, Ni, Sun, Yang, and Chen]{pan2010cross}
Sinno~Jialin Pan, Xiaochuan Ni, Jian-Tao Sun, Qiang Yang, and Zheng Chen.
\newblock Cross-domain sentiment classification via spectral feature alignment.
\newblock In \emph{Proceedings of the 19th international conference on World
  wide web}, pages 751--760, 2010.

\bibitem[Sejdinovic et~al.(2013)Sejdinovic, Sriperumbudur, Gretton, and
  Fukumizu]{Sejd2013equi}
Dino Sejdinovic, Bharath Sriperumbudur, Arthur Gretton, and Kenji Fukumizu.
\newblock Equivalence of distance-based and rkhs-based statistics in hypothesis
  testing.
\newblock \emph{The Annals of Statistics}, 41\penalty0 (5):\penalty0
  2263--2291, 2013.
\newblock ISSN 00905364, 21688966.
\newblock URL \url{http://www.jstor.org/stable/23566550}.

\bibitem[Sun et~al.(2015)Sun, Feng, and Saenko]{sun2015return}
Baochen Sun, Jiashi Feng, and Kate Saenko.
\newblock Return of frustratingly easy domain adaptation, 2015.

\bibitem[Sun et~al.(2019)Sun, Tzeng, Darrell, and Efros]{sun2019unsupervised}
Yu~Sun, Eric Tzeng, Trevor Darrell, and Alexei~A Efros.
\newblock Unsupervised domain adaptation through self-supervision.
\newblock \emph{arXiv preprint arXiv:1909.11825}, 2019.

\bibitem[Wang et~al.(2020)Wang, Shelhamer, Liu, Olshausen, and
  Darrell]{wang2020fully}
Dequan Wang, Evan Shelhamer, Shaoteng Liu, Bruno Olshausen, and Trevor Darrell.
\newblock Fully test-time adaptation by entropy minimization, 2020.

\bibitem[Wei et~al.(2020)Wei, Hu, Weng, Xing, Yu, and Luo]{wei2020learning}
Xiangpeng Wei, Yue Hu, Rongxiang Weng, Luxi Xing, Heng Yu, and Weihua Luo.
\newblock On learning universal representations across languages, 2020.

\bibitem[{Wolf} et~al.(2019){Wolf}, {Debut}, {Sanh}, {Chaumond}, {Delangue},
  {Moi}, {Cistac}, {Rault}, {Louf}, {Funtowicz}, {Davison}, {Shleifer}, {von
  Platen}, {Ma}, {Jernite}, {Plu}, {Xu}, {Le Scao}, {Gugger}, {Drame},
  {Lhoest}, and {Rush}]{2019arXiv191003771W}
Thomas {Wolf}, Lysandre {Debut}, Victor {Sanh}, Julien {Chaumond}, Clement
  {Delangue}, Anthony {Moi}, Pierric {Cistac}, Tim {Rault}, R{\'e}mi {Louf},
  Morgan {Funtowicz}, Joe {Davison}, Sam {Shleifer}, Patrick {von Platen},
  Clara {Ma}, Yacine {Jernite}, Julien {Plu}, Canwen {Xu}, Teven {Le Scao},
  Sylvain {Gugger}, Mariama {Drame}, Quentin {Lhoest}, and Alexander~M. {Rush}.
\newblock {HuggingFace's Transformers: State-of-the-art Natural Language
  Processing}.
\newblock \emph{arXiv e-prints}, art. arXiv:1910.03771, October 2019.

\bibitem[Yang et~al.(2019)Yang, Cheng, Liu, and Sun]{yang-etal-2019-reducing}
Zonghan Yang, Yong Cheng, Yang Liu, and Maosong Sun.
\newblock Reducing word omission errors in neural machine translation: A
  contrastive learning approach.
\newblock In \emph{Proceedings of the 57th Annual Meeting of the Association
  for Computational Linguistics}, pages 6191--6196, Florence, Italy, July 2019.
  Association for Computational Linguistics.
\newblock \doi{10.18653/v1/P19-1623}.
\newblock URL \url{https://www.aclweb.org/anthology/P19-1623}.

\bibitem[Yu and Jiang(2016)]{yujiang2016learning}
Jianfei Yu and Jing Jiang.
\newblock Learning sentence embeddings with auxiliary tasks for cross-domain
  sentiment classification.
\newblock In \emph{Proceedings of the 2016 Conference on Empirical Methods in
  Natural Language Processing}, pages 236--246, Austin, Texas, November 2016.
  Association for Computational Linguistics.
\newblock \doi{10.18653/v1/D16-1023}.
\newblock URL \url{https://www.aclweb.org/anthology/D16-1023}.

\bibitem[Zhang et~al.(2019)Zhang, Zhang, Liu, Zhao, Zhu, and
  Chen]{zhang2019interactive}
Kai Zhang, Hefu Zhang, Qi~Liu, Hongke Zhao, Hengshu Zhu, and Enhong Chen.
\newblock Interactive attention transfer network for cross-domain sentiment
  classification.
\newblock In \emph{Proceedings of the AAAI Conference on Artificial
  Intelligence}, volume~33, pages 5773--5780, 2019.

\bibitem[Zhao et~al.(2019)Zhao, Combes, Zhang, and Gordon]{zhao2019learning}
Han Zhao, Remi Tachet~des Combes, Kun Zhang, and Geoffrey~J Gordon.
\newblock On learning invariant representation for domain adaptation.
\newblock \emph{arXiv preprint arXiv:1901.09453}, 2019.

\bibitem[Ziser and Reichart(2017)]{ziser2017neural}
Yftah Ziser and Roi Reichart.
\newblock Neural structural correspondence learning for domain adaptation,
  2017.

\bibitem[Ziser and Reichart(2018)]{ziser-reichart-2018-pivot}
Yftah Ziser and Roi Reichart.
\newblock Pivot based language modeling for improved neural domain adaptation.
\newblock In \emph{Proceedings of the 2018 Conference of the North {A}merican
  Chapter of the Association for Computational Linguistics: Human Language
  Technologies, Volume 1 (Long Papers)}, pages 1241--1251, New Orleans,
  Louisiana, June 2018. Association for Computational Linguistics.
\newblock \doi{10.18653/v1/N18-1112}.
\newblock URL \url{https://www.aclweb.org/anthology/N18-1112}.

\bibitem[Ziser and Reichart(2019)]{ziser-reichart-2019-task}
Yftah Ziser and Roi Reichart.
\newblock Task refinement learning for improved accuracy and stability of
  unsupervised domain adaptation.
\newblock In \emph{Proceedings of the 57th Annual Meeting of the Association
  for Computational Linguistics}, pages 5895--5906, Florence, Italy, July 2019.
  Association for Computational Linguistics.
\newblock \doi{10.18653/v1/P19-1591}.
\newblock URL \url{https://www.aclweb.org/anthology/P19-1591}.

\end{thebibliography}

\end{document}